# Title: The Right to Be Remembered: Preserving Maximally Truthful Digital Memory in the Age of AI

**Authors:** Alex Zhavoronkov, PhD [1,2,3] [*]; Dominika Wilczok [4,5], Roman Yampolskiy, PhD [6]

[1] Insilico Medicine AI Limited, Masdar, Abu Dhabi, UAE
[2] Insilico Medicine US Inc, Cambridge, MA 02138, USA
[3] Insilico Medicine Hong Kong Ltd., Hong Kong, Hong Kong SAR
[4] Duke University, Durham, NC27708, USA
[5] Duke Kunshan University, Jiangsu, Kunshan China
[6] Speed School of Engineering, University of Louisville, Louisville, KY40292, USA

**Corresponding author**: Alex Zhavoronkov, PhD. alex@insilico.cm

**Abstract:**
Since the rapid expansion of large language models (LLMs), people have begun to rely on them for information retrieval. While traditional search engines display ranked lists of sources shaped by search engine optimization (SEO), advertising, and personalization, LLMs typically provide a synthesized response that feels singular and authoritative. While both approaches carry risks of bias and omission, LLMs may amplify the effect by collapsing multiple perspectives into one answer, reducing users' ability or inclination to compare alternatives. This concentrates power over information in a few LLM vendors whose systems effectively shape what is remembered and what is overlooked. As a result, certain narratives, individuals or groups, may be disproportionately suppressed, while others are disproportionately elevated. Over time, this creates a new threat: the gradual erasure of those with limited digital presence, and the amplification of those already prominent, reshaping collective memory.
To address these concerns, this paper presents a concept of the Right To Be Remembered (RTBR) which encompasses minimizing the risk of AI-driven information omission, embracing the right of fair treatment, while ensuring that the generated content would be maximally truthful.

## Introduction

LLMs are neural architectures trained on immense corpora of text using a self-supervised learning objective: given a sequence of tokens, the model predicts the probability of the next token [1]. These models are almost universally built on the transformer architecture whose self-attention mechanism allows the network to capture both short- and long-range dependencies in text efficiently [2]. The training data include billions of words across diverse domains ranging from scientific papers and literature to news articles and technical manuals or online discussions, allowing the model to internalize a statistical representation of language at multiple levels. The depth and scale of the architecture allow it to store and recombine an extraordinary range of associations, appearing as if the model is "knowledgeable" [3]. However, their knowledge is not stored as discrete facts in a database but as distributed patterns in parameter space that when prompted, reconstruct plausible continuations. Retrieval-augmented generation (RAG) was developed to address the limitations of this emergent but unreliable knowledge by searching large document collections or the live web, identifying relevant passages, and feeding them back into the generative model [4]. The model then conditions its output on this evidence, ideally reducing hallucination and providing verifiable citations. However, if information is absent, misrepresented, or marginalized in the training corpus, the model cannot generate it reliably. RAG, for its part, can only retrieve what is already digitized, indexed, and accessible [5]. The reliability of AI outputs is further undermined by the instability of digital information such as expired links, lapsed domains, obsolete file formats, removed web pages, content moderation takedowns, algorithmic de-ranking, and deliberate flooding of digital spaces with misleading or repetitive content. Empirical studies of "link rot" show that a substantial portion of online content disappears within a few years, with some analyses suggesting that the median URL becomes inaccessible about a decade later [6][7]. For AI systems dependent on both static training datasets and live retrieval, these losses translate directly into informational amnesia, shaping not only what AI systems can produce, but also what users come to believe is available to be known. Users rarely encounter a visible "gap"; instead, they receive a fluent answer that conceals the underlying incompleteness. Over time, these dynamic privileges majority narratives, entrenches existing inequalities of visibility, and risks erasing those whose traces are sparse or fragile. In practical terms, entire scientific contributions may vanish from citation chains, marginalized communities may be further silenced, and public discourse may narrow to what is most digitized and most persistent. The consequences extend beyond technical error, touching on recognition, fairness and epistemic justice in the digital age. Confronting these dynamics demands a framework that treats digital memory not as an incidental by-product of computation but as a site of justice, where fair representation and the preservation of diverse voices are actively safeguarded. This need can be expressed as a right to be remembered (RBTR) which insists on the preservation of human presence and contribution within the digital record.

**Why it matters to be remembered**

The RTBR can be understood as a normative claim that contributions, whether scientific, cultural, or social, should remain accessible within the digital infrastructures that increasingly determine what counts as knowledge. Remembrance is not a neutral by-product of information systems, but a process structured by institutional, economic, and technological forces. RTBR acquires its strongest meaning when understood through the lens of contribution. Human progress depends on the accumulation of knowledge, yet in practice the traces of many contributions are fragile. For instance, scientific papers without digital publications produced in languages other than English, and research outside of prestigious institutions often fall outside the field of retrieval systems that underpin both scholarly databases and AI-mediated knowledge engines. A 2017 large-scale analysis of 496,665 PubMed-indexed biomedical articles from 1966-2015 revealed that only 40.48% overall had Digital Object Identifiers (DOIs), with stark geographic disparities in availability: western countries like the United States and United Kingdom achieved 46-57% overall coverage (peaking at 89-97% by 2015), while nations such as Russia, Ukraine, and Thailand exhibited near-zero adoption (0-2%), rendering pre-2000 publications from these regions particularly vulnerable to erasure in AI-mediated knowledge retrieval [8]. In this way, entire lines of inquiry can effectively vanish, not because they lacked value, but because infrastructures of recognition failed to preserve and amplify them. To be remembered, in this context, is to remain available as part of the foundation upon which future discovery builds.

This principle extends beyond formal achievements to encompass the texture of everyday life. Social history and anthropology have long demonstrated that ordinary practices like domestic work, patterns of consumption, or the rhythms of daily labor are essential to understanding how societies function. In the digital era, where only fragments of ordinary existence are systematically captured, the danger is that our descendants inherit a distorted archive dominated by those who had the resources and incentives to publish and amplify themselves. RTBR also carries a dimension of justice. Patterns of erasure reproduce existing inequalities of visibility. Scholars from the Global South, authors writing in less dominant languages, and women or minority contributors are systematically less represented in digital archives and search outputs[9][10]. Marginalized communities whose histories were already vulnerable to suppression now face an additional layer of exclusion as AI systems overproduce the narratives of the powerful while underproducing those of the less visible [11][12].

Finally, collective memory functions as a public good: it sustains accountability by preserving evidence of wrongdoing, it enables progress by transmitting knowledge across generations, and it fosters fairness by recognizing contributions that might otherwise be excluded. In this sense, the RTBR articulates a societal duty: to maintain a truthful and inclusive record of human experience. It is not merely an individual entitlement to visibility but a collective commitment to safeguarding the plurality of memory itself.

The consequences of neglecting the RTBR reach far beyond the mechanics of information retrieval. Collective memory, rather than being just a record of the past; it is the substrate on which future knowledge is built. Remembrance functions as an enabling condition for intellectual and cultural continuity. To preserve contributions is not only to honor past labor but to safeguard the raw material of future thought. In the sciences, this means that neglected studies or unindexed findings retain the possibility of resurfacing when new methods or theories render them relevant. When contributions vanish from visibility, they foreclose the conditions under which others might have engaged, refined, or challenged them. In this sense, erasure alters the trajectory of inquiry itself, narrowing the possibilities of discovery or possibilities of finding potential fallacies in reasoning, leading to taking false conclusions for granted. In the arts and humanities, it means that practices excluded from the dominant record can continue to inform how communities understand themselves. Across domains, the RTBR affirms that knowledge is cumulative only if its layers remain accessible. What is at stake, then, is not whether AI systems provide correct answers in the present, but whether they participate in maximally truthful preserving the scaffolding of human understanding for the future. The infrastructures of memory built today will determine whose voices endure and whose are silenced in the next century of knowledge. To treat remembrance as a design principle is to recognize that memory is a public good: a shared inheritance that sustains accountability, enables renewal, and ensures that progress does not proceed by way of forgetting.

**Who is responsible for the forgetting**

The responsibility for bias, omission, or distortion of LLM outputs lies primarily with the vendors who design and deploy them, though the structural tendencies of the models themselves play an important role. Vendors exert direct control at multiple stages: they decide which datasets are included in training and which are excluded, thereby shaping the boundaries of what the model can "know" [13]. They implement filtering and moderation policies to align outputs with corporate values [14], cultural norms [15], or regulatory requirements [16], choices that inevitably suppress certain information while amplifying others, such as filtering what the company classifies as "harmful content" which could inadvertently suppress legitimate but controversial perspectives. Vendors also fine-tune models through reinforcement learning with human feedback (RLHF), embedding human judgements into the system in ways that can entrench biases or steer responses toward preferred narratives [17]. Deployment design adds another layer of influence: whether users are presented with a single synthesized answer or multiple perspectives, whether citations are provided, and how much control individuals have over outputs are all decisions made by vendors, not by the models themselves.

At the same time, LLMs themselves are not neutral containers. Recent empirical studies confirm that LLMs do not simply reflect vendor-imposed filters but also exhibit systematic, topic-dependent biases of their own, such as political leanings that persist even across different architectures and tasks [18]. Due to their statistical nature, they amplify dominant patterns in the training corpus while filtering out rare, marginal, or less digitally visible perspectives. Their tendency to simplify, collapse multiple viewpoints, and produce fluent, authoritative-sounding text creates a structural bias toward mainstream or majority narratives, regardless of vendor intent [19]. Moreover, the users also contribute to the bias; the phrasing of prompts can shape responses, potentially eliciting biased or incomplete answers if prompts are vague or leading, though vendors still control the degree of flexibility users have in steering outputs [20].
Thus, the risk of erasure or distortion emerges from an interplay: vendors actively shape memory through their choices and incentives, while models passively but powerfully reinforce existing inequalities in representation by overproducing the common and underproducing the rare. Recognizing this dual responsibility is essential for developing safeguards: while accountability must be demanded from vendors as stewards of design and policy, technical solutions must also address the intrinsic tendencies of LLMs to forget or distort the margins of human knowledge.

**Ensuring maximal truthfulness**

The pursuit of maximal truthfulness in AI demands a layered framework that recognizes truth as simultaneously a statistical property of model outputs, an epistemic relation to external sources of evidence, and a normative requirement of honesty in communication. Accuracy refers to the degree to which model statements correspond to empirical reality, while honesty captures whether the model's responses are consistent with its internal representations rather than strategically distorted. The distinction is now formalized in emerging benchmarks such as the MASK framework, which attempts to measure whether models state what they "believe," even when pressured to conform to misleading prompts [21]. Here, a model's 'belief' means the consistent answer it gives to neutral prompts (e.g., saying Paris is the capital of France; dishonesty is measured when pressured answers contradict that belief, while accuracy is measured by comparing the belief itself to the ground truth.

The internal structure of neural networks also provides signals about truthfulness. Probing studies have identified what has been called a "truth direction" in activation space, a geometric dimension that reliably distinguishes correct from incorrect answers [22]. Stronger models appear to exhibit this direction more consistently, suggesting that truth may become increasingly linearly separable in high-dimensional representation space as scale grows. Complementary research has shown that factually correct and incorrect model outputs differ in the geometry of the model's internal representations. Using local intrinsic dimension analysis, investigators found that truthful responses are encoded in more compact, lower-dimensional activation patterns, whereas hallucinations are scattered across higher-dimensional manifolds showing that truthfulness can sometimes be read directly from a model's internal dynamics, possibly allowing unreliable generations to be flagged even without external labels[23].

How credit is assigned within AI-mediated knowledge systems raises a fundamental tension between efficiency and recognition. For most users, the priority is to obtain concise and actionable information rather than a full genealogy of its production. Yet this streamlining process risks obscuring the people and communities whose work made the answer possible. Over time, contributors may vanish from view, because attribution was collapsed into an anonymous synthesis. Designing systems that preserve recognition without compromising efficiency is therefore crucial. Provenance can be layered so that immediate answers remain accessible while the deeper scaffolding of citations and credit remains retrievable, demonstrated by initiatives such as EKILA for digital art [24], and the C2PA content authenticity standard, where metadata and attribution trails can be embedded directly into outputs. Extending these ideas to language models suggests that efficient information delivery and robust recognition of contributors need not be mutually exclusive; provenance can be layered, preserving the visibility of scientific and cultural labor even as answers are streamlined. However, what is still missing is a systematic way for AI systems to carry forward scholarly citation chains and intellectual labor the way academic publishing does. Current solutions largely capture data lineage and content authenticity, but they don't embed epistemic acknowledgment such as remembering the researchers, labs, and communities whose work underpins the outputs.

Finally, truthfulness is inseparable from transparency about uncertainty. Models must be designed not only to provide answers but also to signal when no reliable answer can be given. Recent experiments with calibrated probabilities of "knowing" versus "not knowing" demonstrate that language models can learn to withhold judgment when evidence is insufficient, thereby avoiding confident falsehoods [25]. Such abstention mechanisms are essential to a conception of maximal truthfulness that prioritizes honesty over fluency.

Taken together, these strands of research indicate that maximal truthfulness requires aligning the internal representations of models with the external record of human knowledge, embedding systematic verification into inference, and cultivating norms of epistemic humility. Progress will depend on weaving together advances in representation analysis, retrieval, attribution, verification, and calibration into a coherent architecture that treats truth not as an incidental by-product but as a design objective.

**Right to Be Remembered vs Right to Erasure**

The proposed RBTR must be contextualized against the established legal and ethical framework commonly known as the "right to be forgotten". Emerging from European jurisprudence, this framework provides individuals with mechanisms to control their digital footprint. While the concept gained initial prominence through the 2014 *Google Spain* decision regarding search engine delisting [26], its scope is defined most significantly by Article 17 of the General Data Protection Regulation (GDPR), officially termed the "Right to Erasure", which mandates that data controllers erase personal data under several circumstances, even if the data is accurate and current [27]. These grounds include situations where the data is no longer necessary for its original purpose, where the individual withdraws consent and no other legal basis for processing exists, where the individual successfully objects to the processing (and no overriding legitimate grounds are demonstrated), or if the data was processed unlawfully [27]. This broad right was conceptualized primarily to manage the risks posed by searchable commercial databases and expansive search indices, providing vital protection for individual autonomy and privacy [28].

In these traditional systems, data is stored discretely, and its deletion or delisting is technically straightforward. In contrast, the LLMs internalize and synthesize the data during training, embedding knowledge within their vast parameter spaces, in the end making information erasure challenging [3]. This characterization is making the LLMs the vaults of human history, as each piece of data represents some part of the human experience, which is aligned with the idea behind the RTBR. However, it puts the RTBR in direct conflict with the Right to Erasure as enabling the removal of accurate data based on withdrawal of consent or individual objection, presents an intractable obstacle to achieving the goal of maximally truthful AI. The interplay of the Right to Erasure and the rise of LLMs creates a new phenomenon: "machine unlearning". It can be defined as the process of extracting internalized data from the model's parameters. This field remains nascent; current methods are computationally expensive, imperfect, and often risk degrading the model's overall utility and coherence [29]. Attempting to excise data from a trained model is akin to performing a "lobotomy" on the repository of human knowledge. Erasing models "memories" at the request of regulators or corporate overseers risks damaging the integrity of the entire system and introducing gaps and distortions into the historical record.

We argue that in the context of foundational AI, the collective RTBR including ensuring maximal truthfulness and historical accuracy must take precedence over individual claims to erasure. While the GDPR recognizes exceptions to erasure for freedom of expression and historical research, the societal imperative to maintain a complete and undistorted history of humanity within these powerful systems generally outweighs the broad privacy mandates established for commercial data processing. This argument gains particular salience concerning individuals who are deceased or dying. While privacy interests generally fade after death, the historical and societal value of their life experiences remains [30]. It is essential that the entire digital legacy of the deceased be actively preserved, encoded, summarized, and integrated into foundational models. This ensures that the models shaping our future understanding of the past are based on the most complete data available, safeguarding a nuanced historical record for future generations.

**Discussion**

The concept of the RBTR proposed in this work reframes the preservation of digital memory as an ethical, technical, and societal imperative in the age of AI. While the Right to Be Forgotten was established to protect individual autonomy and privacy in a world of persistent digital traces, the current information ecosystem presents the inverse challenge: an accelerating erosion of records and contributions that form the basis of collective knowledge. In this new context, forgetting is no longer the default risk of neglect but the engineered outcome of biased curation, corporate filtering, and algorithmic omission. LLMs, now positioned as the de facto mediators of digital knowledge, not only retrieve but also reconstruct information in a manner that profoundly influences what is remembered. Their statistical architectures transform memory from a discrete archival process into a dynamic act of synthesis, which is deeply contingent on the visibility, accessibility, and stability of the underlying data. The absence of evidence thus becomes evidence of absence, and over time, the erosion of digital traces translates into epistemic inequity favoring those with sustained digital presence and institutional reinforcement.

Preserving maximal truthfulness, therefore, cannot rely solely on technical correction mechanisms such as retrieval-augmentation or post-hoc fact-checking. Truthfulness must be reimagined as a multi-level construct encompassing accuracy, provenance, honesty, and inclusivity. Empirical evidence from geometric analyses of model activations suggests that truth may indeed have a structural signature within neural networks, a finding that invites future research into architectures designed to align internal representations with verifiable knowledge states.

This demands a shift in AI design calling for systems that explicitly optimize for epistemic integrity. Mechanisms for layered attribution, traceable provenance, and uncertainty calibration must become as fundamental to AI architectures as token prediction or reinforcement alignment. Similarly, regulatory frameworks should evolve to reflect the dual necessity of protecting personal privacy while safeguarding the completeness of collective memory. Current legislation, including the GDPR, was conceived in the context of static databases, not adaptive knowledge engines that encode vast portions of humanity's intellectual history. As foundational models increasingly shape the epistemic foundations of future generations, prioritizing the RTBR alongside privacy rights will be essential to sustaining the integrity of shared knowledge.

Finally, remembrance in this context must be seen not as the perpetuation of all data, but as the preservation of meaningful, contextually grounded human contribution. The challenge is to ensure that the archive of human thought remains as inclusive, accurate, and representative as possible.

**Conclusion**

As AI systems become the dominant interface to human knowledge, the integrity of what they remember and what they omit will increasingly define the boundaries of collective understanding. RTBR articulates a new ethical foundation for this era: that digital memory should be preserved as a public good, and that truthfulness in AI depends not only on epistemic precision but on the inclusivity and continuity of the human record from which it learns. To preserve maximally truthful digital memory is to preserve the conditions of human understanding itself. RTBR presents multiple philosophical questions on equal and fair treatment of individual life data, achievements, ratings, ranking, and the impact of digital immortality on human behavior and decision making. However, the authors argue that the RTBR and maximal truthfulness are net positive on both individual and population levels and facilitate optimal human-AI convergence.

**Authors' contributions:** AZ - original idea and writing. DW -ideas, writing and editing. RY - ideas and writing.

**Conflicts of Interests:** The authors declare no conflicts of interests